\documentclass[pdflatex,sn-mathphys]{sn-jnl}% Math and Physical Sciences Reference Style
\jyear{2021}%

\theoremstyle{thmstyleone}%
\theoremstyle{thmstyletwo}%

\theoremstyle{thmstylethree}%
\def\In{{\mathrm{in}}}
\def\Out{{\mathrm{out}}}
\def\Conv{{\mathrm{conv}}}
\def\Synapse{{\mathrm{synapse}}}
\def\Mfo{{\mathrm{mfo}}}
\def\Adder{{\mathrm{adder}}}
\def\Euclid{{\mathrm{euclid}}}
\def\Cos{{\mathrm{cos}}}
\def\R{{\mathbb{R}}}

\DeclareMathOperator{\sign}{sign}
\usepackage{pifont}% http://ctan.org/pkg/pifont
\newcommand{\cmark}{\ding{51}}%
\newcommand{\xmark}{\ding{55}}%
\usepackage{soul}

\def\t{{^\top}}
\def\x{{\mathbf{x}}}
\def\w{{\mathbf{w}}}

\newcommand{\norm}[1]{\left\lVert#1\right\rVert_2^2 }
\newcommand{\rootnorm}[1]{\left\lVert#1\right\rVert_2 }

\begin{document}

\title[EuclidNets]{EuclidNets: An Alternative Operation for Efficient Inference of Deep Learning Models}

%%=============================================================%%
%% Prefix	-> \pfx{Dr}
%% GivenName	-> \fnm{Joergen W.}
%% Particle	-> \spfx{van der} -> surname prefix
%% FamilyName	-> \sur{Ploeg}
%% Suffix	-> \sfx{IV}
%% NatureName	-> \tanm{Poet Laureate} -> Title after name
%% Degrees	-> \dgr{MSc, PhD}
%% \author*[1,2]{\pfx{Dr} \fnm{Joergen W.} \spfx{van der} \sur{Ploeg} \sfx{IV} \tanm{Poet Laureate} 
%%                 \dgr{MSc, PhD}}\email{iauthor@gmail.com}
%%=============================================================%%

\author[1]{\fnm{Xinlin} \sur{Li}}\email{xinlin.li1@huawei.com}

\author[2]{\fnm{Mariana} \sur{Parazeres}}\email{moprazeres@outlook.com}
%\equalcont{These authors contributed equally to this work.}

\author[2]{\fnm{Adam} \sur{Oberman}}\email{adam.oberman@mcgill.ca}
%\equalcont{These authors contributed equally to this work.}

\author[1]{\fnm{Alireza} \sur{Ghaffari}}\email{alireza.ghaffari@huawei.com}

\author[2]{\fnm{Masoud} \sur{Asgharian}}\email{masoud.asgharian2@mcgill.ca}

\author*[1,3]{\fnm{Vahid} \sur{Partovi~Nia}}\email{vahid.partovinia@huawei.com}

\affil*[1]{\orgdiv{Huawei Noah's Ark Lab}, \orgname{Montreal Research Centre}, \orgaddress{\street{7101 Park Avenue}, \city{Montreal}, \postcode{H3N 1X9}, \state{QC}, \country{Canada}}}

\affil[2]{\orgdiv{Department of Mathematics and Statistics}, \orgname{McGill University}, \orgaddress{\street{805 Sherbrooke Street West}, \city{Montreal}, \postcode{H3A 0B9}, \state{QC}, \country{Canada}}}

\affil[3]{\orgdiv{Department of Mathematics and Industrial Engineering}, \orgname{Polytechnique Montreal}, \orgaddress{\street{2500 Chem. Polytechnique}, \city{Montreal}, \postcode{H3T 1J4}, \state{QC}, \country{Canada}}}

%%==================================%%
%% sample for unstructured abstract %%
%%==================================%%

\abstract{With the advent of deep learning application on edge devices, researchers actively try to optimize their deployments on low-power and restricted memory devices. There are established compression method such as quantization, pruning, and architecture search that leverage commodity hardware. Apart from conventional compression algorithms, one may redesign the operations of deep learning models that lead to more efficient implementation. To this end, we propose EuclidNet, a compression method, designed to be implemented on hardware which replaces multiplication, $xw$, with Euclidean distance $(x-w)^2$.  
We show that EuclidNet is aligned with matrix multiplication and it can be used as a measure of similarity in case of convolutional layers. Furthermore, we show that under various transformations and noise scenarios, EuclidNet exhibits the same performance compared to the deep learning models designed with multiplication operations.
}

%%================================%%
%% Sample for structured abstract %%
%%================================%%

% \abstract{\textbf{Purpose:} With the advent of deep learning application on edge devices, researchers actively try to optimized their deployments on low-power and restricted memory devices. There are established compression method such as quantization, pruning, and architecture search are designed for commodity hardware. On the other hand, apart from conventional compression algorithms, there is a possibility to redesign the operations of deep learning models that lead to more efficient implementation.

% \textbf{Methods:} We propose EuclidNet, a compression method, designed to be implemented on hardware which replaces multiplication, $xw$, with Euclidean distance $(x-w)^2$.  EuclidNet allows for a low precision hardware implementation which is about twice as efficient (in term of logic gate counts) as the comparable conventional hardware, with negligible loss of accuracy. 

% \textbf{Results:} We show theoretically that EuclidNet is aligned with matrix multiplication and it can be used as a measure of similarity in case of convolutional layers. Furthermore, we show that under various transformations and noise scenarios, EuclidNet shows the same performance compared to the networks designed with multiplication operations.

% \textbf{Conclusion:} Euclidean distance is a prominent method in replacing expensive multiplication operations in hardware. EuclidNet performs as well as multiplicative deep learning models under different test scenarios
% }

\keywords{Deep learning compression, Euclidean distance, Convolutional Neural Network, Hardware efficient algorithms}

%%\pacs[JEL Classification]{D8, H51}

%%\pacs[MSC Classification]{35A01, 65L10, 65L12, 65L20, 65L70}

\maketitle

\section{Introduction}\label{sec1}

While majority of deep neural networks are trained on GPUs, they are increasingly being deployed on edge devices, such as mobile devices. These edge devices require 
%compressed (more efficient)  \emph{hardware aware} architectures 
to compress the architecture for a given hardware design (e.g. GPU or lower precision chips)  due to memory and power constraints \cite{benmeziane2021comprehensive, cheng2017survey}. Moreover, application specific hardware are being designed to accommodate the deployment of deep learning models. Thus,  designing efficient deep learning architectures that are efficient for the deployment (i.e. \emph{inference}) has become a new challenge in the deep learning community.
%This leads to a new problem formulation which we address here: \emph{design an efficient hardware architecture which allows networks to be trained on GPUs, then implemented on the hardware.}

The combined problem of hardware and deep learning model design is complex, and the precise measurement of efficiency is both device and model specific. This is because researchers have to  take into account various efficiency factors such as latency, memory footprint, energy consumption.  Here we deliberately oversimplify the problem in order to make it tractable, by addressing a fundamental element of hardware cost. Knowing that power consumption is directly related to the chip area in a digital circuit, we use the chip area required to implement an arithmetic operation on a hardware as a surrogate to measure the efficiency of a deep learning model. While this is very coarse, and full costs will depend on other aspects of hardware implementation, it nevertheless represents a fundamental unit of cost in hardware design \cite{hennessy2011computer}.

\begin{figure}
    \centering
    \includegraphics[width=0.32\textwidth]{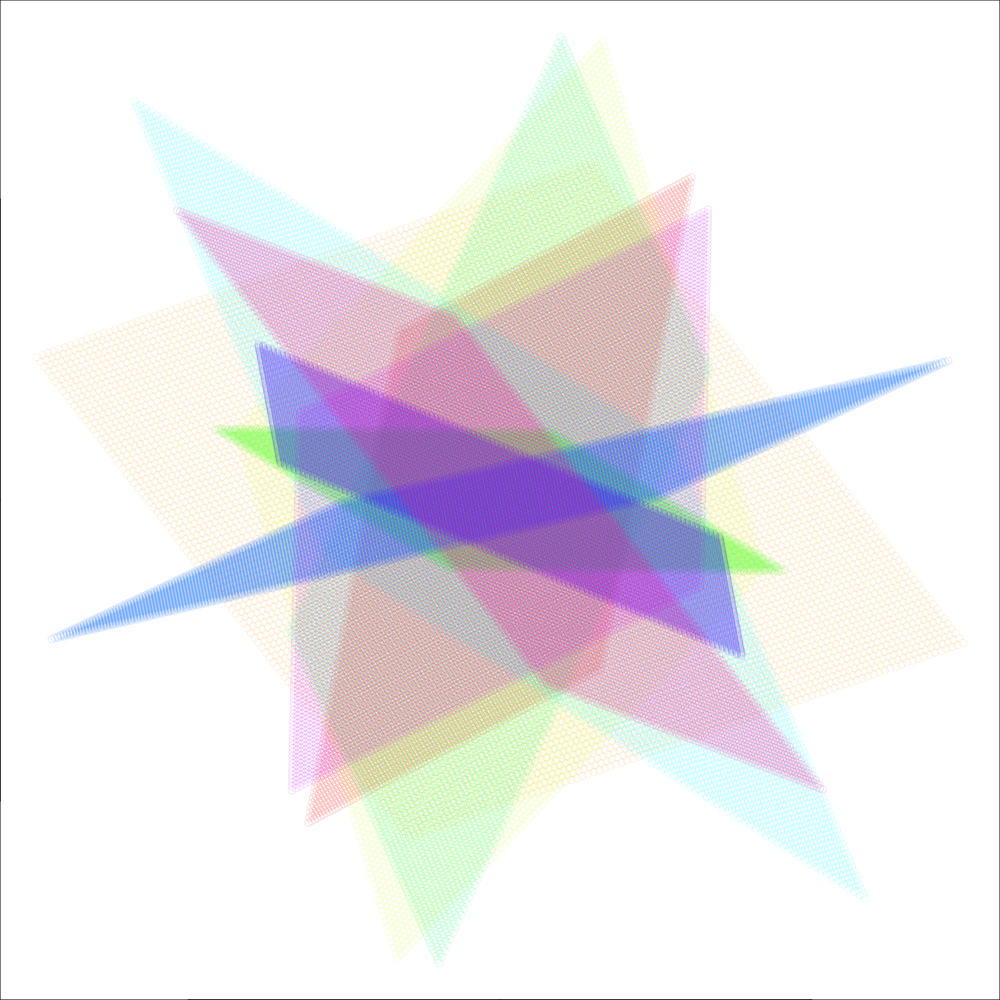}
    \includegraphics[width=0.32\textwidth]{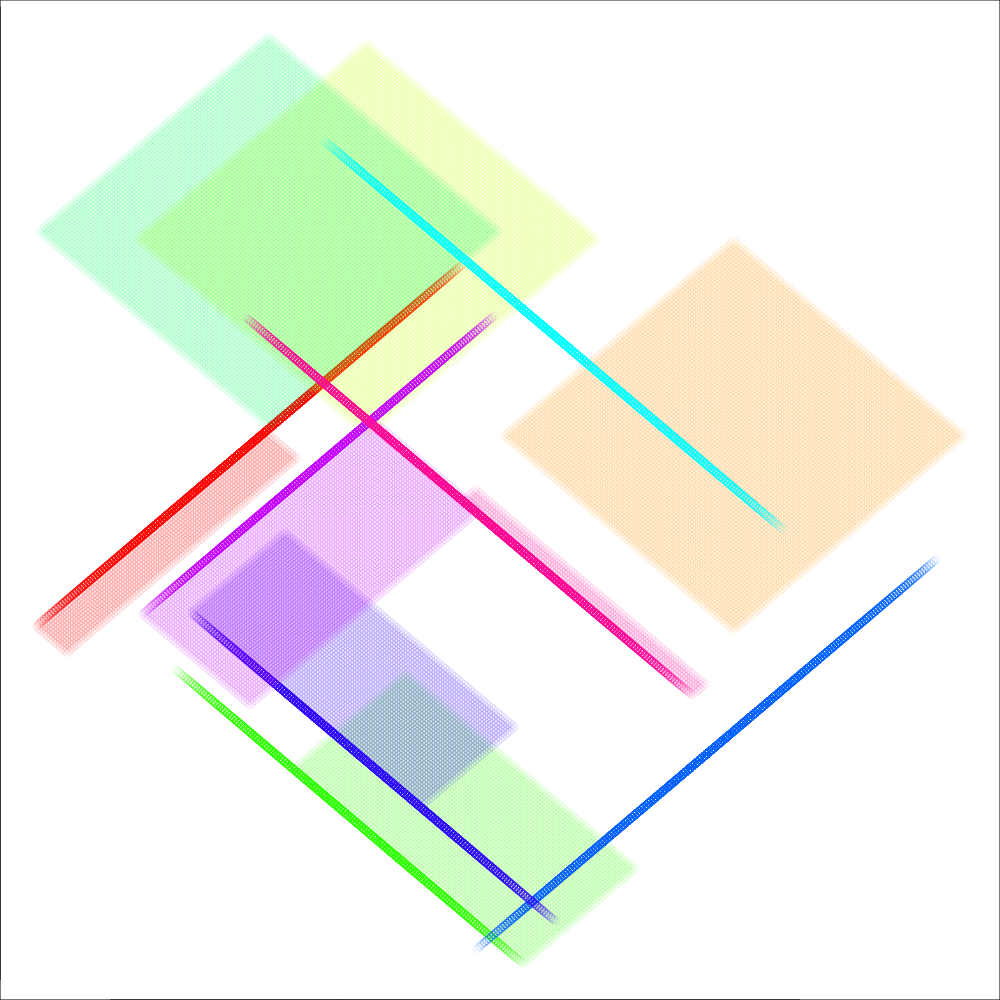}
    \includegraphics[width=0.32\textwidth]{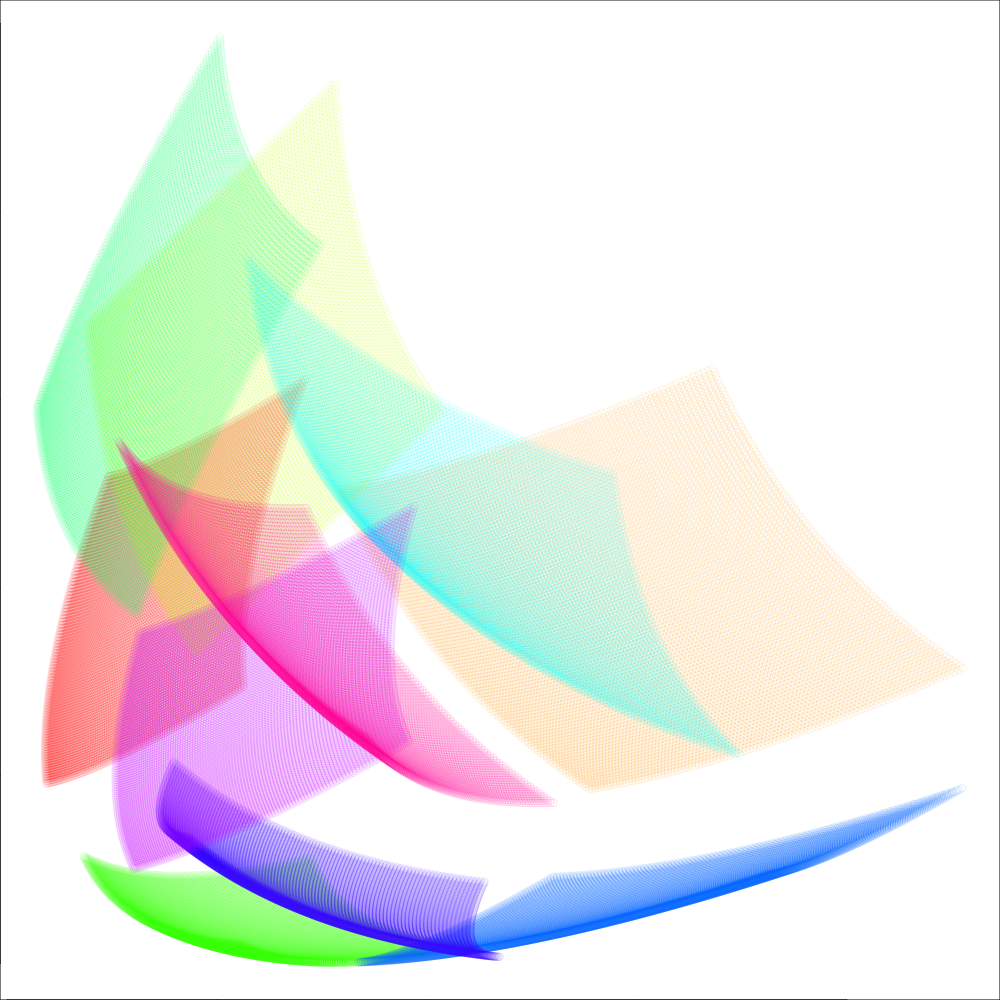}
    \caption{Feature representation of traditional convolution with $S(x,w) = xw$ (left), AdderNet $S(x,w) = -\|x-w\|_1$ (middle), EuclidNet $S(x,w) = -\frac{1}{2}\|x-w\|_2^2$ (right).} 
    \label{fig:feature}
\end{figure}

In a deep learning model, weights are multiplied by inputs, hence on of the fundamental operations in deep learning models is multiplication $S_{\Conv}(x,w) = wx$. In our work, we replace multiplication with the EuclidNet operator,
\begin{equation}\label{eq: euclid}
    S_{\Euclid}(x,w) = -\frac{1}{2}\|x-w\|_2^2.
\end{equation}
which combines a difference with a square operator. We will refer to the family of deep learning models that use equation \eqref{eq: euclid} as EuclidNets. 
These models compromise between standard multiplicative models and AdderNets\cite{chen2020addernet}, which remove multiplication entirely, but at the cost of a significant loss of accuracy and difficult training procedure. Replacing multiplication with square can potentially reduce the computation cost. The feature representation of each of the architectures is illustrated in Figure~\ref{fig:feature}. EuclidNets can be implemented on 8-bit precision without loss of accuracy as demonstrated in Table~\ref{tab: quant}. 

 The square operator is cheaper than multiplication and it can also be implemented using look up tables \cite{de2009large}.  In \cite{baluja2018no,covell2019table}, authors prove that replacing look up table can replace actual float computing, while works such as LookNN in \cite{razlighi2017looknn} take the first step in designing hardware for look up table use. On a low precision hardware, we can compute $S_\Euclid$ for about half the cost of computing $S_\Conv$.
 %, because hardware efficiencies for squaring two a fixed precision integer more than offsets the additional cost of a difference.  
 Furthermore, using EuclidNets,  the deep learning model does not lose expressivity, as explained ins Section \ref{sec:theory}. 
To summarize, we make the following contributions:
\begin{itemize}
    \item We design a deep learning architecture based on replacing the multiplication $S_\Conv(x,w) = wx$ by the squared difference equation \eqref{eq: euclid}.  We show that using square operator can potentialy reduce the hardwaer cost. 
    \item These deep learning models are just as expressive as convolutional neural networks.  In practice, they have comparable accuracy (drop of less than 1 percent on ImageNet on ResNet50 going from full precision convolutional to 8-bit EuclidNets).
    \item We show theoretically and empirically that EuclidNets have the same behaviour compared to convolutional neural network in the case that the input is transformed (e.g. linear transformation) or affected by noise (e.g. Guassian noise).
    \item  We provide an easy approach to train EuclidNets using homotopy.
\end{itemize}

\begin{table}[h]
	\caption{Euclid-Net Accuracy with full precision and 8-bit quantization: Results on ResNet-20 with Euclidian similarity for CIFAR10 and CIFAR100, and results on ResNet-18 for ImageNet.  Euclid-Net achieves comparable or better accuracy with 8-bit precision, compared to the conventional full precision convolutional neural network. 
	}\label{tab: quant}
	\centering

\begin{tabular}{ccc ccc}
    % table head
	
	\multirow{3}{*}{\textbf{Network}} &
	\multirow{3}{*}{\textbf{Quantization}} &
	\multirow{3}{*}{\textbf{Chip Efficiency}} &
%	\multicolumn{2}{c}{\textbf{Accuracy}} \\ 
%	&&
%	\multirow{2}{*}{\textbf{CIFAR10}} &
%	\multirow{2}{*}{\textbf{CIFAR100}} &
	\multicolumn{3}{c}{\textbf{Top-1 accuracy}} \\ 
	&&& CIFAR10 & CIFAR100& ImageNet \\
	\hline
	% content
	\multirow{2}{*}{$S_{\Conv}$} &
	Full precision &	\xmark &	92.97%92.25 
	& 68.14 & 69.56 \\%69.8 \\ 
 		&	8-bit &	\cmark &	92.07 & 68.02 &	69.59 \\
		\multirow{2}{*}{$S_{\Euclid}$} &
	Full precision &    \xmark & 93.32 & 68.84 & 69.69 \\ 
		&	8-bit &	\cmark &	93.30 & 68.78 & 68.59 \\
		\multirow{2}{*}{$S_{\Adder}$} &
	Full precision &    \xmark & 91.84 & 67.60 & 67.0 \\ 
		&	8-bit &	\cmark &	91.78 & 67.60 & 68.8 \\
			\multirow{1}{*}{BNN} &
	1-bit &    \cmark & 84.87 & 54.14 & 51.2 \\
\end{tabular}

\end{table}

 \section{Context and related work}

Compressing deep learning models comes at the costs of accuracy loss, and increasing training time (to a greater extent on quantized networks) \cite{frankle2018lottery, cheng2018model}. 
Part of the accuracy loss comes simply from decreasing model size, which is required for mobile and edge devices  \cite{wu2019machine}. Some of the most common deep learning compression methods include pruning \cite{reed1993pruning}, quantization \cite{guo2018survey}, knowledge distillation \cite{hinton2015distilling}, and efficient design \cite{iandola2016squeezenet,howard2017mobilenets,zhang2018shufflenet,tan2019efficientnet}. Between the compression methods, the most prominent approach is low bit quantization \cite{guo2018survey}. In this case, the inference can speed up with lowering bit size, at the cost of 
 accuracy drop and  longer training times. In the extreme quantization, such as binary networks, operations have negligible cost at inference but exhibits a considerable accuracy drop \cite{Hubara_BNN}.

Here we focus on a small sub-field of compression, that optimizes mathematical operations in a deep learning model.  This approach can be combined successfully with other conventional compression methods, such as quantization \cite{xu2020kernel} and pruning \cite{reed1993pruning}.

 On the other hand, knowledge distillation \cite{hinton2015distilling} consists of transferring information from a larger teacher network to a smaller student network. The idea is easily extended by thinking of information transfer between different similarity measures, which \cite{xu2020kernel} explores in the context of AdderNets. Knowledge distillation is an uncommon training procedure and requires extra implementation effort. However, EuclidNet preserves the accuracy without knowledge distillation. We suggest a straightforward training using a smooth transition between common convolution and Euclid operation using homotopy.

%\section{Background}

\section{Similarity and Distances}

\subsection{Inner Products versus Distances}

Consider an intermediate layer of a deep learning model  with input $x\in\R^{H\times W \times c_{\In}}$ and output $y~\in~\R^{H\times W \times c_{\Out}}$ where $H,W$ are the dimensions of the input feature, and $c_{\In}, c_{\Out}$ the number of input and output channels, respectively. For a standard convolutional network, we represent the input to output transformation via weights $w~\in~\R^{d\times d\times c_{\In}\times c_{\Out}}$ as
\begin{equation}\label{eq: layer}
y_{mnl} = \sum_{i = m}^{m+d} \sum_{j=n}^{n+d} \sum_{k = 0}^{c_{\In}} 
x_{ijk} w_{ijkl}
%S(x_{ijk}, w_{ijkl}).
\end{equation}
Setting $d=1$ reduces the equation \eqref{eq: layer} to a fully-connected layer.
We can abstract the multiplication of the weights $w_{ijkl}$ by $x_{ijkl}$ in the equation above by using a similarity measure $S:\R\times\R\to\R$. The original convolutional layer corresponds to 
$$
S_{\Conv}(x,w) = xw.
$$
In our work, we replace $S_\Conv$ with $S_\Euclid$, given by equation \eqref{eq: euclid}.  A number of works have also replaced the multiplication operator in  deep learning models. 
The most relevant work is the AdderNet \cite{chen2020addernet}, which uses 
\begin{equation}\label{eq: adder}
S_{\Adder}(x,w) = -\|x-w\|_1.
\end{equation}
to replace multiplication by $\ell_1$ norm, i.e. summation of the absolute value of differences. 
This operation can be implemented very efficiently on a custom hardware, knowing that subtraction and absolute value of different $n$-bit integers cost $\mathcal{O}(n)$ gate operations, compared to $\mathcal{O}(n^2)$ for multiplication i.e. $S_\Conv(x,w) = xw$. However, AdderNet comes with a significant loss in accuracy, and is difficult to train. 

\subsection{Other Similarity Measures}
%We start by introducing other similarities already studied in the literature that are closely related to our work. Then, we give a short overview of compression methods and explore how they fit within a single  framework.

%\subsection{ } \label{sec: litrev}

The idea of replacing multiplication operations to save resources within the context of neural networks dates back to 1990s. Equally motivated by computational speed-up and hardware logic minimization, authors of \cite{dogaru1999comparative} defined perceptrons that use the synapse similarity,
\begin{equation}\label{eq: comp_syn}
S_{\Synapse}(x,w) = \sign(x)\cdot \sign(w) \cdot \min(\|x\|,\|w\|),
\end{equation}
which is cheaper than multiplication in terms of hardware complexity.

Equation \eqref{eq: comp_syn} has not been experimented with modern deep learning models and datasets. Moreover, in \cite{akbas2015multiplication} a slight variation is introduced which is also a multiplication-free operator,
\begin{equation}\label{eq: mf}
S_{\Mfo}(x,w) = \sign(x)\cdot\sign(w)\cdot(\|x\|+\|w\|)).
\end{equation}
Note that both equations \eqref{eq: comp_syn} and \eqref{eq: mf} use $\ell_1$-norm. Also note that in \cite{mallah2018multiplication}, the updated design choice  allows contributions from both operands $x$ and $w$. Furthermore, in \cite{afrasiyabi2018non},  the similarity in image classification on CIFAR10 is studied. Other applications of  equation \eqref{eq: mf} are studied in \cite{badawi2017multiplication, pan2019additive}.

In \cite{you2020shiftaddnet}, the similarity operation is further combined with a bit-shift, leading to an improved accuracy with negligible added hardware cost. However, the accuracy results for AdderNet appear to be lower than those reported in \cite{chen2020addernet}. Another follow-up work uses knowledge distillation to further improve the accuracy of AdderNets \citep{xu2020kernel}. 

Instead of simply replacing the similarity on the summation, there is also the possibility to replace the full expression of equation \eqref{eq: layer} as, for example,  proposed in  \cite{limonova2020resnet,limonova2020bipolar}, by approximating the activation of a given layer with an exponential term. Unfortunately, these methods only lead to speed-up in certain cases and, in particular, they do not improve CPU inference time. Moreover, the reported accuracy on the benchmark problems is also lower than the typical baseline.

In \cite{mondal2019dense}, authors used three layer morphological neural networks for image classification. Morphological neural networks were introduced in  1990s in \cite{davidson1990theory, ritter1996introduction} and use the notion of erosion and dilation to replace equation \eqref{eq: layer}:
 \begin{align*}
 \mbox{Erosion}(x,w) &= \min_j S(x_j, w_j) = \min_j (x_j - w_j), \\
 \mbox{Dilation}(x,w) &= \max_j S(x_j, w_j) = \max_j (x_j + w_j).
 \end{align*}
The authors proposed two methods by stacking layers to expand networks, but they admitted the  possibility of  over-fitting and difficult training issues, casting doubt on scalability of the method.

\section{Theoretical Justification}\label{sec:theory}
This section provides some theoretical ground for the connections among AdderNets, EuclidNets, and conventional convolution.
\subsection{Equivalence with Multiplication}\label{sec:theory_align}

Euclidean distance has a close tie with multiplication and hence, it can replace the multiplications in convolution and linear layers. Here, we delve into the details of this claim a bit more.

Let us consider Euclidean distance between the two vectors $\x$ and $\w$ as $\norm{\x-\w}=(\x-\w)\t(\x-\w)$ where $\x$ and $\w$ are the vectors of inputs and weights respectively. Moreover, $\x$ and $\w$ are vectors of random variables, so it is of interest to study the expected value of the EuclidNet operation first,
\begin{equation}
    -\frac 1 2 \mathbb{E} \norm{\x-\w} = -\frac 1 2 \mathbb{E}\norm \x - \frac 1 2 \mathbb{E}\norm \w + \mathbb{E}(\x\t\w) .
    \label{eq:euclid_expect}
\end{equation}
In other words \eqref{eq:euclid_expect}, convlution similarity measure, i.e.  the inner product $\x\t\w$, is embedded in EuclidNet form. However, the result is biased with two extra terms i.e. $-\frac 1 2 \mathbb{E}\norm\x$ and $-\frac 1 2 \mathbb{E}\norm\w$. Thus we may  conclude that Euclidean distance is aligned with multiplication shifted by two bias terms. The induced bias by the EuclidNet operation remains controlled in both training or inference, most deep learning models use some sort of normalization mechanism such as batch norm,  layer norm, and weight norm. 

Euclidean distance also has a close relationship with cosine similarity. Let us define $S_\Cos$ as 

\begin{equation}
    S_\Cos(\x,\w):= \frac{\x^\top \w}{\rootnorm \x \rootnorm\w}.
    \label{eq:cosine_sim}
\end{equation}
It is easy to see that in the case of having a normalization mechanism (i.e. $\rootnorm \x=\rootnorm \w=1$) the cosine similarity and Euclid similarity become equivalent
\begin{eqnarray}
    S_\Euclid(\x,\w)= S_\Cos(\x,\w)-1 &\mathrm{\quad s.t.}& \rootnorm \x = \rootnorm \w=1 .
    \label{eq:cosine_distance}
\end{eqnarray}

Moreover, Euclidean norm is a transitive similarity measure since it satisfies the following inequality %(inverse triangle inequality)

\begin{equation}
    \|\x-\w\|_2 \geq\lvert~\|\x\|_2-\|\w\|_2 ~\rvert .
    \label{eq:inv_triangle}
\end{equation}
It is noteworthy to mention that this transitivity holds for p-norms (i.e. $\|\mathbf{a}\|_p= (\sum_i \|{a}_i\|^p)^{\frac{1}{p}}$). This means that the AdderNet \cite{chen2020addernet} operator is also transitive. According to equation \eqref{eq:cosine_distance}, however, the only norm that has such a close relationship with the cosine similarity is Euclidean norm. This is the distinguishing feature of the EuclidNets that while they are distance based, and hence enjoy the transitivity property in measuring similarity, their performance is also completely aligned with those based on Cosine similarity.

% Although batch-norm and layer-norm can compensate the EuclidNet bias, we are still eager to understand the extent of this effect. We can use Cauchy–Schwarz inequality

% \begin{equation}
%     \|X.W\|^2 \leq \|X\|^2\|W\|^2 .
%     \label{eq:cs_ineq}
% \end{equation}

% Furthermore, since $W$ and $X$ are bounded, $a \leq W_i \leq A$ and $b \leq X_i \leq B$, from inverse Cauchy–Schwarz inequality we have

% \begin{equation}
%     \|X\|^2\|W\|^2  \frac{4}{\left (\sqrt{\frac{ab}{AB}}+\sqrt{\frac{AB}{ab}} \right )^2}\leq \|X.W\|^2  .
%     \label{eq:inv_cs_ineq}
% \end{equation}

% By combining inequalities \eqref{eq:cs_ineq} and \eqref{eq:inv_cs_ineq} we can conclude

% \begin{equation}
%     \frac{4}{\left (\sqrt{\frac{ab}{AB}}+\sqrt{\frac{AB}{ab}} \right )^2}\leq \frac{\|X.W\|^2}{\|X\|^2\|W\|^2} \leq 1  .
%     \label{eq:inv_cs_comb}
% \end{equation}

\subsection{Expressiveness of EuclidNets}
Deep learning models that use the EuclidNet operation are just as expressive as those using multiplication. Note the polarization identity, 
\[
S_\Conv(x,w) = S_\Euclid(x,w) - S_\Euclid(x,0) - S_\Euclid(0,w)
\]
which means that any multiplication operation can be expressed using only Euclid operations.  

\subsection{Hardware cost}

Traditionally, hardware developers 
use smaller multipliers to create larger multipliers \cite{de2009large}. They use various methods of multiplier tiling or divide and conquer to form larger multiplier. Karatsuba algorithm and its generalization \cite{weimerskirch2006generalizations} is among the most known algorithms to implement large multipliers. Here we show that Euclidean distance can be potentially implemented with fewer multipliers in hardware.

Karatsuba algorithm is a form of divide and conquer algorithm to perform $n-$bit multiplication using $m-$bit multipliers. Let us assume $a$ and $b$ are $n-$bit integer numbers and they can be re-written using two $m-$bits partitions
\begin{align}
\nonumber
    &a = a_1 \times 2^m + a_2,\\ \nonumber
    &b = b_1 \times 2^m + b_2.\\ 
    \label{eq:karatsuba_parts}
\end{align}
In the case of multiplication, we have
\begin{align}
\nonumber
    &ab = (a_1 \times 2^m + a_2) (b_1 \times 2^m + b_2)\\ \nonumber
    &~~~= 2^{2m}a_1b_1+2^m a_1b_2+2^m a_2b_1+a_2b_2,\\ 
    \label{eq:mult}
\end{align}
which is comprised of \textit{three} additions and \textit{four} $m-$bits multiplications. However for the squaring operation, we have
\begin{align}
\nonumber
    &a^2 = (a_1 \times 2^m + a_2) (a_1 \times 2^m + a_2)\\ \nonumber
    &~~~= 2^{2m}a_1^2+2^{m+1} a_1a_2+a_2^2,\\
    \label{eq:square}
\end{align}
which is comprised of \textit{two} additions and \textit{three} $m-$bits multiplications. Thus, the squaring operation can be cheaper in hardware. Also note that such divide and conquer techniques are used commonly in designing accelerator on FPGA targets.

\section{Training EuclidNets}

Training EuclidNets are much easier compared to other similarity measures such as AdderNets. This makes EuclidNet attractive for complex tasks such as image segmentation, and object detection where training compressed networks are challenging and causes large accuracy drops. However, EuclidNets are more expensive than AdderNets when using floating-point number format, however, their quantization is easy since, unlike AdderNets, they behave  similar to traditional convolution to a great extent. In another words EuclidNets are easy to quantize.

While training a deep learning model using EuclidNets, it is more appropriate to use the identity 
\begin{equation}
    S_{\Euclid}(x,w) = -\frac {x^2}{2} - \frac{w^2}{2} + x w,
\end{equation}  
that is more appropriate for GPUs that are optimized for inner product computations. As such,  training EuclidNets does not require additional CUDA kernel implementation unlike AdderNets \citep{cuda}. The official implementation of AdderNet \citep{chen2020addernet} reflects order of  $20\times$ slower training than the traditional convolution on PyTorch. This is specially problematic for large deep learning models and complex tasks since even traditional convolution training takes few days or even weeks. EuclidNet training is about $2\times$ slower in the worst case and their implementation is natural in deep learning frameworks such as PyTorch and Tensorflow.

\begin{table}[h]\label{tab: times}
	\caption{Time (seconds) and maximum training batch-size that can fit in a signle GPU \textit{Tesla V100-SXM2-32GB}, during ImageNet training. In parenthesis is the slowdown with respect to the $S_{conv}$ baseline. 
	We do not show times for AdderNet, which is much slower than both, because it is not implemented in CUDA
	}
	\centering
\begin{tabular}{cc l l cc}
	\multirow{2}{*}{\textbf{Model}} & \multirow{2}{*}{\textbf{Method}} & \multicolumn{2}{l}{\textbf{ Maximum Batch-size}} & \multicolumn{2}{l}{\textbf{Time per step}} \\ 
	&  & \multicolumn{1}{l}{\textbf{\begin{tabular}[c]{@{}c@{}} power of 2\end{tabular}}} & \multicolumn{1}{l}{\textbf{integer}} & \textbf{Training} & \textbf{Testing} \\ \hline
	\multirow{2}{*}{ResNet-18} & $S_{\Conv}$ & 1024 & 1439 & 0.149 & 0.066 \\ 
	& $S_{\Euclid}$ & 512 & 869 ($1.7\times$) & 0.157 ($1.1\times$) & 0.133 ($2\times$) \\ \hline
	\multirow{2}{*}{ResNet-50} & $S_{\Conv}$ & 256 & 371 & 0.182 & 0.145 \\ 
	&  $S_{\Euclid}$ & 128 & 248 ($1.5\times$) & 0.274 ($1.5\times$) & 0.160 ($1.1\times$) \\ \hline
\end{tabular}

\end{table}

A common method in training neural networks is fine-tuning, that means initializing with weights trained on different data but with a similar nature. Here, we introduce the idea of using a weight initialization from a model trained on a related similarity measure. 

Rather than training from scratch, we wish to fine-tune EuclidNet starting from accurate  CNN weights.  This is achieved by an ``architecture homotopy" where we change hyperparameters to convert a regular convolution to an EuclidNet operation 
\begin{equation}
S(x,w; \lambda_k) = xw - \lambda_k\frac{x^2 + w^2}{2},\qquad \mbox{ with }\lambda_k = \lambda_0 + \frac{1 - \lambda_0}{n} \cdot k,
\label{eq: homotopy}
\end{equation} 
where $n$ is the total number of epochs and $0<\lambda_0<1$ is the initial  transition phase. Note that $S(x,w,0) = S_{\Conv}(x,w)$ and $S(x,w,1) =S_{\Euclid}(x,w)$ and equation \eqref{eq: homotopy} is a convex combination of these two similarities. One may interpret  $\lambda_k$ as a scheduler for the homotopy, similar to the way learning rate is scheduled in training a deep learning model. We found that a linear scheduling as shown in equation \eqref{eq: homotopy} is empirically effective.

Transformations like equation \eqref{eq: homotopy} are commonly used in scientific computing \cite{allgower2003introduction}. The idea of using homotopy  in training neural networks can be traced back to \cite{chow1991homotopy}. Recently, homotopy was used in deep learning in the context of activation functions \citep{pathak2019parameter,cao2017hashnet, mobahi2016training,farhadi2020}, loss functions \citep{gulcehre2016mollifying}, compression \citep{chen2019efficient} and  transfer learning \citep{bengio2009curriculum}. Here, we use homotopy in the context of transforming operations of a deep learning model.

Fine-tuning method in equation \eqref{eq: homotopy} is inspired by continuation methods in partial differential equations. Assume $S$ is a solution to a differential equation with the initial condition  $S(x,0) = S_0(x)$. In certain situations, solving this differential equation for $S(x,t)$ and then evaluating at $t=1$ might be easier than solving directly for $S_1$. One may think of this homotopy method as an evolution for deep learning model weights. At time zero the deep learning model consists of regular convolutional layers, but they gradually transform to Euclidean layers.

\begin{figure}[t]
	\begin{center}
	\includegraphics[width=0.7\linewidth]{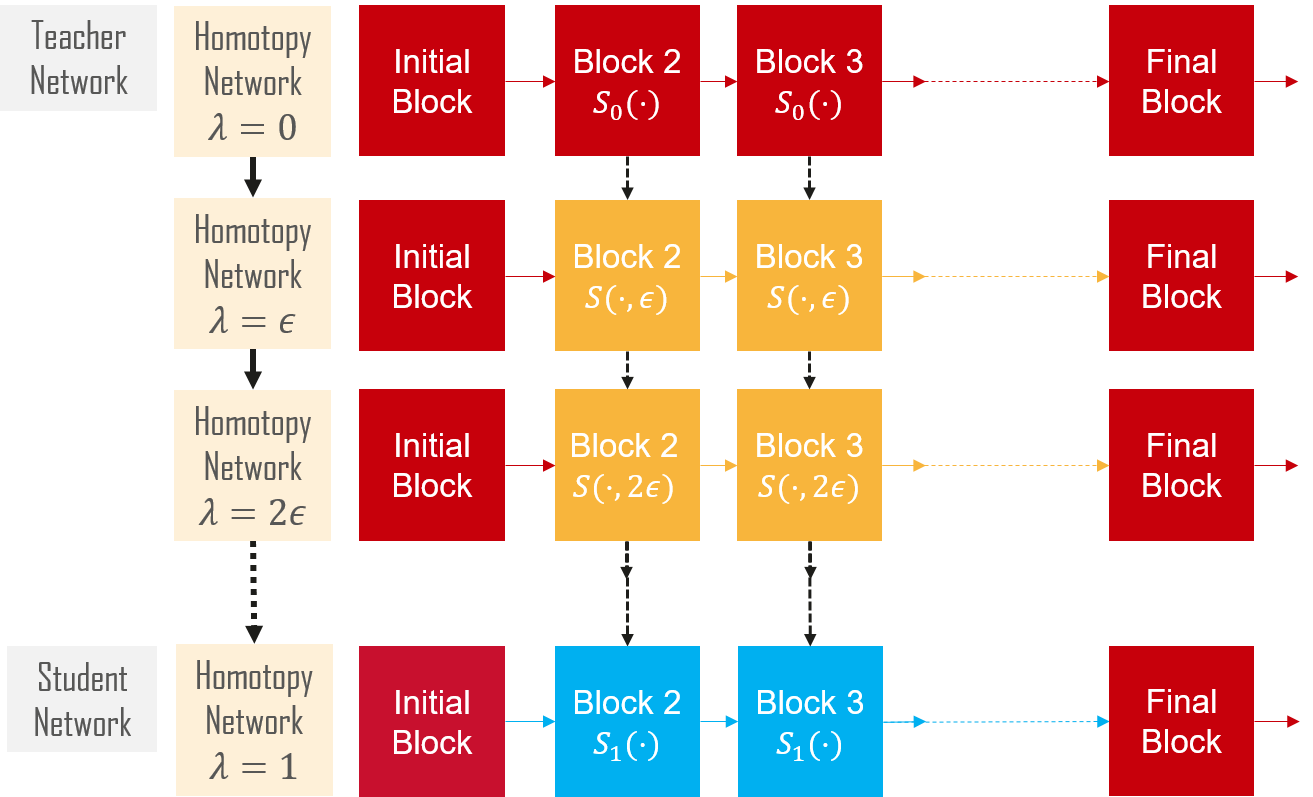}
	\end{center}
	\caption{Training schema of EuclidNet using Homotopy, i.e. transitioning from traditional convolution $S(x,w)=xw$ towards EuclidNet $S(x,w)=-\frac{1}{2} |x-w|^2$ through equation \eqref{eq: homotopy}.} \label{fig: homotopy}
\end{figure}

The homotopy method can also be interpreted as a sort of  of knowledge distillation. Whereas knowledge distillation methods tries to match a student network to a teacher network, the homotopy can be seen as a slow transformation from the teacher network into a student network. Figure \ref{fig: homotopy}  demonstrates the idea. Interestingly, problems that have been solved with homotopy have also been tackled by knowledge distillation \citep{hinton2015distilling,chen2019efficient,yim2017gift, bengio2009curriculum}. %For example, removing blocks or layers from a deep learning model \citep{hinton2015distilling,chen2019efficient,} along with transfer learning \citep{yim2017gift, bengio2009curriculum}. 

\section{Experiments}\label{sec:Experiments}

To illustrate performance of the EuclidNets, We apply our proposed method on image classification tasks. We also test our trained deep learning model under different transformations on the input image and compare the accuracy to standard convolutional networks.

\subsection{CIFAR10}\label{sec: cifar10}

First, we consider the CIFAR10 dataset, consisting of $32\times32$ RGB images with 10 possible classifications \citep{krizhevsky2009learning}. We normalize and augment the dataset with random crop and random horizontal flip. We consider two ResNet models \cite{he2015deep}, ResNet-20 and ResNet-32.

We train EuclidNet using the optimizer from \cite{chen2020addernet}, which we will refer to as AdderSGD, to evaluate EuclidNet under a similar setup. We use initial learning rate $0.1$ with cosine decay, momentum $0.9$, batch size 128 and weight decay $5\times 10^{-4}$. We follow \cite{chen2020addernet} in setting the learning-rate scaling parameter $\eta$. For traditional convlutional network, we use the same hyper-parameters with stochastic gradient descent optimizer.

The details of classification accuracy is provided in Table \ref{tab: cifar10}. We consider two different weight initialization for EuclidNets. First, we initialize the weights randomly and second, we initialize them with pre-trained on a convolutional network. The accuracy for EuclidNets has negligible accuracy loss compared to the standard ResNets. We see that for CIFAR10 training from scratch achieves even a higher accuracy, while initializing with convolution network and using linear homotopy training improves it even further.

\begin{table}[h]
\caption{Results on CIFAR10. The initial learning rate is adjusted for non-random initialization. 
%{\vahid Do we need to report results on AdderNet to show we beat it in terms of accuracy.}  
%{\adam we mostly need these in the first table}
}
\label{tab: cifar10}
\centering
		\begin{tabular}{ccccccc}
			\multirow{2}{*}{Model} & \multirow{2}{*}{Similarity} & \multirow{2}{*}{Initialization} & \multirow{2}{*}{Homotopy} & \multirow{2}{*}{Epochs} & \multicolumn{2}{c}{Top-1 accuracy} \\ 
			&  &  &  &  & CIFAR10 & CIFAR100 \\ \hline
			\multirow{4}{*}{ResNet-20} & $S_{\Conv}$ & Random & None & 400 & 92.97 & \textbf{69.29} \\ 
			& \multirow{3}{*}{$S_{\Euclid}$} & Random & None & 450 & {93.00} & 68.84 \\ 
			&  & \multirow{2}{*}{Conv} & None & 100 & 90.45 & 64.62 \\ 
			&  &  & Linear & 100 & \textbf{93.32} %92.25 
			& 68.84%66.52 
			\\ \hline
			\multirow{4}{*}{ResNet-32} & $S_{\Conv}$ & Random & None & 400 & \textbf{ 93.93} & 71.07 \\ 
			& \multirow{3}{*}{$S_{\Euclid}$} & Random & None & 450 & 93.28 & \textbf{71.22} \\ 
			&  & \multirow{2}{*}{Conv} & None & 150 & 91.28 & 66.58 \\ 
			&  &  & Linear & 100 & 92.62 & 68.42 \\ \hline
		\end{tabular}

\end{table}

\begin{figure}	\centering
	\includegraphics[width=0.7\textwidth]{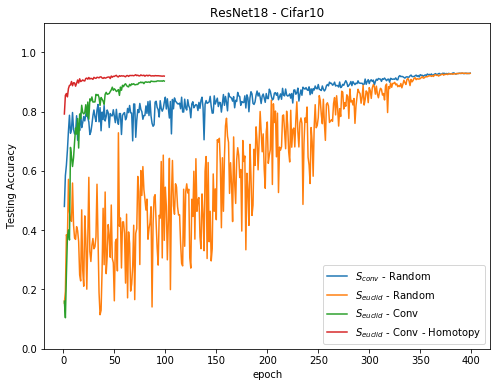}
	\caption{Evolution of testing accuracy during training of ResNet-20 on CIFAR10, initialized with random weights, or initialized from convolution pre-trained network. Initializing from a pretrained convolution network speeds up the convergence. EuclidNet is harder to train compared with convolution network when both initialized from random weights.}\label{fig: train_comparison}
\end{figure}

EuclidNets can become unstable during the training, despite careful choice of the  optimizer. Figure \ref{fig: train_comparison} shows a comparison of the EuclidNet training with a standard convolutional network.  As it can be seen in the Figure \ref{fig: train_comparison}, fine-tuning the EuclidNets directly from convolutional networks' weights is more stable than training from scratch. Also observe that when we train EuclidNets from scrach, accuracy is lower but the convergence is faster. Finally, using homotopy in the training procedure, the accuracy is improved.  Note that the pre-trained convolution weights are commonly available in the most of neural compression tasks, so initializing EuclidNets with pre-trained convolution is a commonplace procedure in optimizing deep learning models for inference.

%\subsubsection{Non-euclidean similarities}
EuclidNets are not only faster to train compared to other norm based similarity measures, but also stand superior in terms of accuracy. AdderNet performs slightly worse in terms of accuracy and also is much slower to train. The accuracy is significantly lower for the synapse \cite{dogaru1999comparative} and the multiplication-free \cite{akbas2015multiplication} operators. Table \ref{tab: sim_comparison} demonstrates a top-1 accuracy comparison of different methods. The reported results on AdderNet are from \cite{xu2020kernel}. Note that although for AdderNet in \cite{xu2020kernel}, authors  used knowledge distillation to close the gap with the full precision, it still falls short compared with EuclidNet. 
%Other competitors $S_{\Synapse}$ and $S_{\Mfo}$ are far worse, see Figure \ref{fig: sim_comparison_plots} in the Appendix for more details.

\begin{table}
	\caption{Full precision results on ResNet-20 for CIFAR10 for different multiplication-free similarities.}
	\label{tab: sim_comparison}
	\centering
	\begin{tabular}{ c  c c c c c}
		\multirow{1}{4em}{\textbf{Similarity}} &  \multirow{1}{3em}{$S_{\Conv}$} & \multirow{1}{3em}{$S_{\Euclid}$} & $S_{\Adder}$  & $S_{\Mfo}$ & $S_{\Synapse}$  \\ 
%		& & & \cite{xu2020kernel} &\cite{akbas2015multiplication} & \cite{dogaru1999comparative}\\
		\hline
		\textbf{Accuracy} & 92.97 %93.11 
		& \textbf{93.00}% \textbf{93.32} 
		& 91.84 %%92.96%%
		& 82.05 & 73.08		 
		\\ 
	\end{tabular}
\end{table}

%\subsubsection{ Quantization in CIFAR10}\label{sec: quant}

Training a quantized $S_{\Euclid}$  is very similar to convolutional neural networks. This allows a wider use of such models for lower resource devices. Quantization of the EuclidNets to 8bits keeps the accuracy drop within the range of one percent \citep{wu2020integer} similar to traditional convolutional neural networks.  Table \ref{tab: quant} shows 8-bit quantization of EuclidNet where the accuracy drop remains negligible. Furthermore, training EuclidNets on CIFAR100 dataset exhibits a negligible accuracy drop when the weights are initialized with pre-trained standard model weights.

\subsection{ImageNet}

Next, we consider testing EuclidNet classifier on ImageNet \cite{imagenet_cvpr09} which is known to be a challenging classification task comparing to CIFAR10. We trained our baseline convolutional neural network with standard augmentations of random resized crop and horizontal flip and normalization. We consider ResNet-18 and ResNet-50 models with the same hyper-parameters as those used in Section \ref{sec: cifar10}.

Table \ref{tab: in} shows top-1 and top-5 classification accuracy of ImageNet dataset. As shown in Table \ref{tab: in},  the accuracy of EuclidNet when it is trained from scratch is lower than the baseline emphasizing the importance of homotopy training. We believe that the accuracy drop with no homotopy is because the hyper-parameter tuning is harder for large datasets such as ImageNet. This means that even though there exists hyper-parameters that achieve equivalent accuracy with random initialization, however it is too difficult to find them. Thus, it is much easier to use the existing hyper-parameters of traditional convolutional neural network, and use homotopy to smoothly transfer the weights to wights that are suitable for EuclidNets. 

\begin{table}[h]
	\centering
		\caption{Full precision results on ImageNet. Best result for each model is in bold.}\label{tab: in}
	\scalebox{0.8}{
	\begin{tabular}{ccccccc}
		Model & Similarity & Initialization & Homotopy & Epochs & \multicolumn{1}{l}{Top-1 Accuracy} & \multicolumn{1}{l}{Top-5 Accuracy} \\ \hline
		\multirow{6}{*}{ResNet-18} & $S_{\Conv}$ & Random & None & 90 & 69.56 & 89.09 \\ \cline{2-7} 
		& \multirow{5}{*}{$S_{\Euclid}$} & Random & None & 90 & 64.93 & 86.46 \\ \cline{3-7} 
		&  & \multirow{4}{*}{Conv} & None & 90 & 68.52 & 88.79 \\ \cline{4-7} 
		&  &  & \multirow{3}{*}{Linear} & 10 & 65.36 & 86.71 \\  
		&  &  &  & 60 & 69.21 & 89.13 \\ 
		&  &  &  & 90 & \textbf{ 69.69} & \textbf{ 89.38} \\ \hline
		\multirow{6}{*}{ResNet-50} & $S_{\Conv}$ & Random & None & 90 & 75.49 & 92.51 \\ \cline{2-7} 
		& \multirow{5}{*}{$S_{\Euclid}$} & Random & None & 90 & 37.89 & 63.99 \\ \cline{3-7} 
		&  & \multirow{4}{*}{Conv} & None & 90 & 75.12 & 92.50 \\ \cline{4-7} 
		&  &  & \multirow{3}{*}{Linear} & 10 & 70.66 & 90.10 \\ 
		&  &  &  & 60 & 74.93 & 92.52 \\ 
		&  &  &  & 90 & \textbf{ 75.64} & \textbf{ 92.86} \\ \hline
	\end{tabular}
	}
\end{table}

\subsection{Transformation and blurring}

Here we provide empirical evidence that Euclidean norm is aligned with the multiplication. First, we show that EuclidNets perform as well as standard convolutional neural networks in the case of \textit{pixel transform}. Second, we show that when the image is blurred with Guassian noise, EuclidNets closely follow the behaviour of the convolutional neural networks.

\subsubsection{Pixel transformation}
We define pixel transformation of an image as
\begin{equation}
\mathbf{I_T} = a\mathbf{I}+b,
\label{eq:transform}
\end{equation} 
where $\mathbf{I}$ is a tensor representing the original image, scalars $a$ and $b$ are transformation parameters, and $\mathbf{I_T}$ is the transformed image. Note that in \eqref{eq:transform}, $a$ controls the contrast and $b$ controls the brightness of the image. Such transformations are widely used in various stages of the imaging systems for instance in color correction, and gain-control (ISO).

\begin{figure}[t]
    \centering
    \includegraphics[width=0.45\textwidth]{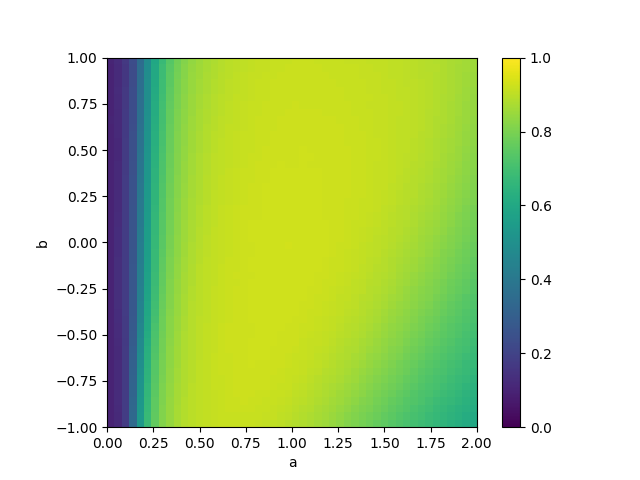}
    \includegraphics[width=0.45\textwidth]{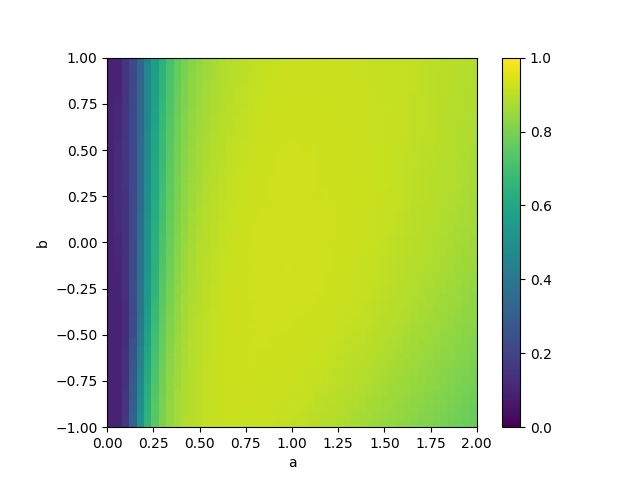}
    \caption{Accuracy of CIFAR10 classification affected by pixel transformation for a standard ResNet-18 (left) and EuclidNet ResNet-18 (right).} 
    \label{fig:transform}
\end{figure}

Figure \ref{fig:transform} shows the accuracy of the standard ResNet-18 and EuclidNet ResNet-18 when the input image is affected by the pixel transformation of equation \eqref{eq:transform}. We can see that when changing $a$ and $b$, EuclidNet ResNet-18 closely follow the accuracy of the standard ResNet-18.

\begin{figure}[t]
    \centering
    \includegraphics[width=0.45\textwidth]{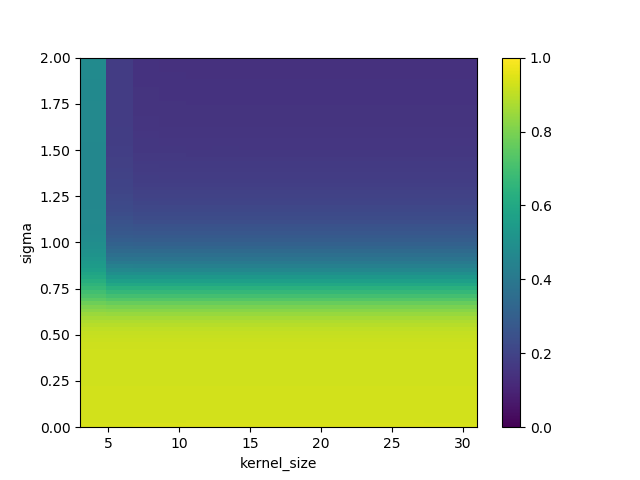}
    \includegraphics[width=0.45\textwidth]{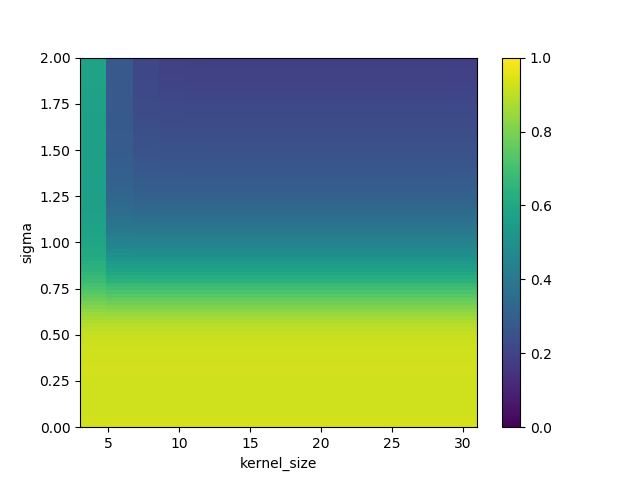}
    \caption{Accuracy of CIFAR10 classification affected by Guassian noise for a standard ResNet-18 (left) and EuclidNet ResNet-18 (right).} 
    \label{fig:noise}
\end{figure}

\subsubsection{Gaussian Blurring}

Additive noise can be injected to an image in different stages of the imaging system due to faulty equipments or environmental conditions. We tested EuclidNet when the input image is affected by a Guassian additive noise. Figure \ref{fig:noise} demonstrates comparison of the standard ResNet-18 and EuclidNet ResNet-18 for different noise intensities i.e. $\sigma$ and kernel sizes. This experiment is done for classification of the CIFAR10 dataset. We can see that EuclidNet ResNet-18 closely follow the behaviour of the standard ResNet-18 in the case of different kernel sizes and noise intensities.

% Hi Alireza, I think the simplest yet most reliable definition and reference would be the one in R. Szeliski's book:
% You can download the book for free here: http://szeliski.org/Book/

% See page 111, Sec. 3.1.1 "Pixel transforms" 
% He defines the formulation as "multiplication and addition with a constant" Eq.(3.3)
% whose parameters control contrast and brightness

% well, this manipulation of contrast and brightness can be part of various stages of an imaging system; e.g., color correction, gain-control (ISO), etc

\section{Conclusion}

EuclidNets are a class of deep learning models in which the multiplication operator is replaced  with the Euclidean distance. They are designed to be implemented on application specific hardware, with the idea that subtraction and squaring are cheaper than multiplication when designing efficient hardware for inference. 
Furthermore, in contrast to other efficient architectures that are difficult to train in low precision, EuclideNets are easily trained in low precision. 
 EuclidNets can be initialized with pre-trained weights of the standard convolutional neural networks and hence, the training procedure of EuclidNets using homotopy is considered as a fine tuning of convolutional networks for inference. The homotopy method further improves training in such scenarios and training using this method sometimes surpass regular convolution accuracy. .

\begin{appendices}

\end{appendices}

%%===========================================================================================%%
%% If you are submitting to one of the Nature Portfolio journals, using the eJP submission   %%
%% system, please include the references within the manuscript file itself. You may do this  %%
%% by copying the reference list from your .bbl file, paste it into the main manuscript .tex %%
%% file, and delete the associated \verb+\bibliography+ commands.                            %%
%%===========================================================================================%%

\bibliography{sn-article.bib}% common bib file
%% if required, the content of .bbl file can be included here once bbl is generated
%%\input sn-article.bbl

%% Default %%
%%\input sn-sample-bib.tex%

\end{document}